\documentclass[runningheads]{llncs}
\usepackage[T2A,T1]{fontenc}
\usepackage[utf8]{inputenc}
\usepackage[russian,finnish,english]{babel}
\usepackage{multirow}
\usepackage{graphicx}[realboxes]
\usepackage{adjustbox}
\usepackage{tipa}
\begin{document}
\title{When Word Embeddings Become Endangered}
%
%
\author{Khalid Alnajjar\orcidID{0000-0002-7986-2994}}
\authorrunning{K. Alnajjar}
%
\institute{Department of Digital Humanities, \\ Faculty of Arts, University of Helsinki, Finland \\
\email{khalid.alnajjar@helsinki.fi}}
\maketitle              
\begin{abstract}
Big languages such as English and Finnish have many natural language processing (NLP) resources and models, but this is not the case for low-resourced and endangered languages as such resources are so scarce despite the great advantages they would provide for the language communities. The most common types of resources available for low-resourced and endangered languages are translation dictionaries and universal dependencies. In this paper, we present a method for constructing word embeddings for endangered languages using existing word embeddings of different resource-rich languages and the translation dictionaries of resource-poor languages. Thereafter, the embeddings are fine-tuned using the sentences in the universal dependencies and aligned to match the semantic spaces of the big languages; resulting in cross-lingual embeddings. The endangered languages we work with here are Erzya, Moksha, Komi-Zyrian and Skolt Sami. Furthermore, we build a universal sentiment analysis model for all the languages that are part of this study, whether endangered or not, by utilizing cross-lingual word embeddings. The evaluation conducted shows that our word embeddings for endangered languages are well-aligned with the resource-rich languages, and they are suitable for training task-specific models as demonstrated by our sentiment analysis model which achieved a high accuracy. All our cross-lingual word embeddings and the sentiment analysis model have been released openly via an easy-to-use Python library.

\end{abstract}

\section{Introduction}
The interest in building natural language processing (NLP) solutions for low-resourced languages is constantly increasing \cite{ws-2019-use}, not only because of the challenges associated with dealing with scarce resources but also because NLP solutions facilitate documenting and analysing languages. Examples of such solutions are applying optical character recognition to scan books \cite{silfverberg2015can}, normalizing historical variation \cite{bollmann2019large}, using speech recognition \cite{hjortnaes2020towards} and more. However, most of the existing research is conducted in a simulated setting \cite{gu2018universal,khayrallah-etal-2020-simulated,126be25f4261423faaf3b829f458ba19} where a reduced portion of the resource-rich language is used to represent a low-resourced language. Other approaches consider Wikipedias of languages having a small number of articles (i.e., $<$ 500,000) such as Latin, Hindi, Thai and Swahili \cite{das2017named,zhou-etal-2020-improving-candidate}.

In this paper, we are dealing with languages that are classified as endangered based on UNESCO Atlas\footnote{http://www.unesco.org/languages-atlas/index.php}. These languages are Erzya\footnote{See \cite{rueter2013erzya} for an insightful description of the situation of the language.} (myv), Moksha (mdf), Komi-Zyrian (kpv) and Skolt Sami (sms). The most common methodology for documenting endangered languages is constructing translation dictionaries, whether digitizing physical dictionaries or reaching to native speakers. Universal dependencies (UD) written by dedicated researchers studying such endangered languages might also be available, and, in a fortunate scenario, they would include translations to a language with more speakers. The bigger languages that endangered languages are translated to are very inconsistent and vary depending on the language family, geographically close languages and the languages spoken by the documenter. 

English, without a doubt, is currently the most resourced language in the field of NLP. However, English translations are not frequently found for endangered and low-resourced languages. To overcome this and make using existing English resources possible, we leverage the recent advances in the field of NLP for aligning word embeddings of big languages such as Finnish and Russian with English word embeddings.

The contributions in this paper are:

\begin{itemize}
    \item Proposing a method for constructing word embeddings for low-resourced and endangered languages, which are also aligned with word embeddings of big languages.
    \item Building a universal sentiment analysis model that achieves high accuracy in both endangered and resource-rich languages covered in this work.
    \item Releasing an open-source and easy-to-use Python library with all the word embeddings and the sentiment analyzer model to support the community and researchers\footnote{\url{https://github.com/mokha/semantics}}.
\end{itemize}

This paper is structured as follows. Section~\ref{sec:rel} contains a brief description of the related work on building cross-lingual low-resource word embeddings. Thereafter, we describe the linguistic resources used in this work, including the translation dictionaries, universal dependencies and existing word embeddings of resource-rich. The proposed method for constructing cross-lingual word embeddings for endangered languages is elaborated then, followed by the description of the sentiment analysis model. We then present the results and evaluation for word embeddings and sentiment analysis model. Lastly, we discuss and highlight our remarks in the conclusions.

\section{Related work}
\label{sec:rel}

The largest scale model for capturing the computational semantics of endangered Uralic languages, Erzya, Moksha, Komi-Zyrian and Skolt Sami, is, perhaps, SemUr \cite{hamalainen2018extracting}. The database consists of words that are connected to each other based on their syntactic co-occurrences in a large internet corpus for Finnish. The extracted relations have been automatically translated by using Jack Rueter's XML dictionaries. In human evaluation, the quality was surprisingly acceptable given that the method was based on word-level translations. This gives hope in using these high-quality dictionaries in building computational semantic models.

Apart from SemUr, there has not been any other attempts in automatically modelling semantics for endangered Uralic languages. Some recent work, however presents interesting work on higher-resourced languages using word embeddings \cite{adams-etal-2017-cross,duong-EtAl2016EMNLP}. In general, word embeddings based methods such as word2vec \cite{mikolov2013efficient} and fastText \cite{bojanowski2017enriching} are optimal for the task of applying high-resource language data to endangered languages as they work on word-level.

Several recent approaches such as GPT-2 \cite{radford2019language}, ELMo \cite{Peters:2018} and BERT \cite{devlin-etal-2019-bert} aim to capture richer semantic representations from text. However, they are very data intensive and their representation is no longer on the level of individual words. This makes it more difficult to use them for endangered languages.


Recently, neural networks have been used heavily in the field of NLP due to their great capabilities in learning a generalization, which resulted in high accuracies. However, neural networks demand a large amount of data, which usually is not available for low-resource languages. Despite this, researchers have employed neural networks in a low-resource setting by producing synthetic data. For instance, Hämäläinen and Rueter have built a neural network to detect cognates for between two endangered languages~\cite{hamalainen-rueter-2019-finding}, Skolt Sami and North Sami. Their approach reached to a better accuracy when they combined data, synthetically produced by a statistical model, with real data.

\section{Linguistic resources}
Here, we describe the linguistic resources used throughout the research presented in this paper. We will focus on resources related to the endangered languages (i.e., Erzya, Moksha, Komi-Zyrian and Skolt Sami), while still providing a brief introduction to resource-rich resources. The resources for endangered languages that we cover here are: 1) translation dictionaries, 2) universal dependencies and 3) finite-state transducers. This list by no means is inclusive of all available and useful resources for endangered languages, as additional resources might exist such as the work of Jack Rueter on Online dictionaries \cite{263590f5c0614385a3fe982ba43fd84e} and making them usable even through click-in-text interfaces \cite{rueter2017giellatekno}. In terms of resource-languages, we describe their word-embeddings.

\subsection{Translation dictionaries}
Low-resource and endangered languages commonly have translation dictionaries to a bigger language. For our case, such dictionaries are multilingual and are provided in an Extensible Markup Language (XML) format. Fortunately, the target languages of the translations are mostly consistent in all the dictionaries (which is not the typical case), but each dictionary contains different portions of translations.

Table~\ref{tab:xml-summary} shows a statistical summary of the translations existing in the dictionaries. The source language represents the endangered languages and the target language indicates the resource-rice language. A meaning group in the dictionaries may contain multiple translations that can be used interchangeably as they share the same meaning. The analysis shows that for Erzya (myv) and Skolt Sami (sms), Finnish (fin) translations are the most common ones, whereas Russian (rus) and English (eng) translations are the most frequent ones for Komi-Zyrian (kpv) and Moksha (mdf).

\begin{table}
\caption{An overview of translation in the XML dictionaries of the low-resourced languages. Language codes are given in ISO 639-3.}\label{tab:xml-summary}
\centering
\begin{tabular}{|c|c|c|c|c|}
\hline
\textbf{Source} & \textbf{Target} & \textbf{Meaning} & \multirow{2}{*}{\textbf{Translations}} & \multirow{2}{*}{\textbf{Total}} \\
\textbf{language} & \textbf{language} & \textbf{groups} & & \\
\hline
\multirow{3}{*}{myv} & fin & \textbf{8388}  & \textbf{14344 (59.89\%)} & \multirow{3}{*}{23950} \\ \cline{2-4} 
                     & rus & 5631           & 7608 (31.77\%)     & \\ \cline{2-4} 
                     & eng & 1917           & 1998 (8.34\%)          & \\ \hline
\multirow{3}{*}{kpv} & fin & 1352           & 2046 (14.89\%)          & \multirow{3}{*}{13744} \\ \cline{2-4} 
                     & rus & \textbf{15492} & \textbf{10585 (77.01\%)} & \\ \cline{2-4} 
                     & eng & 1078           & 1113 (8.10\%)          & \\ \hline
\multirow{3}{*}{sms} & fin & \textbf{20503} & \textbf{27522 (68.02\%)} & \multirow{3}{*}{40461} \\ \cline{2-4} 
                     & rus & 4872           & 6026 (14.89\%)          & \\ \cline{2-4} 
                     & eng & 5824           & 6913 (17.09\%)          & \\ \hline
\multirow{2}{*}{mdf} & rus & 37             & 37 (0.27\%)            & \multirow{2}{*}{13626} \\ \cline{2-4} 
                     & eng & \textbf{6587}  & \textbf{13589 (99.73\%)} & \\ \hline
\end{tabular}
\end{table}

Entries in the dictionaries are in the lemma form, and, typically, their part-of-speech tags are provided. Further metadata information might exist, such as stems and example usages of the word in the source language. We use the Giella \cite{Moshagen2014} dictionaries that have been mainly authored by Jack Rueter through UralicNLP~\cite{uralicnlp_2019}. While Moksha has Finnish translations, the Moksha dictionary in UralicNLP did not contain any of these translations because the data was missing from the repository.

\subsection{Universal dependencies}
Universal dependencies (UD) \cite{11234/1-3424} is a standard framework for annotating the grammar (parts of speech, morphological features, and syntactic dependencies of sentences. Additionally, UD allows annotators to supply their own comments. In the UD we are dealing with, translation sentences might appear in the comments. The UD of the endangered languages can be obtained directly from Universal Dependencies' website\footnote{\url{https://universaldependencies.org/}}. At the time of writing, 1,690, 167, 104 and 435 sentences were in Erzya's~\cite{rueter-tyers-2018-towards}, Moksha's~\cite{mokshaud}, Skolt Sami's~\cite{skoltud} and Komi-Zyrian's UDs\footnote{There is a UD for Komi-Permyak~\cite{rueter-etal-2020-questions} which is close to Komi-Zyrian.}~\cite{partanen-etal-2018-first}, respectively. These numbers highlight the insufficient amount of data present for training machine learning or NLP models for endangered languages.
We have used the UralicNLP~\cite{uralicnlp_2019}, a Python library, to read the universal dependencies.

\subsection{Finite-state transducers}
The common automatic tools found for endangered languages are finite-state transducers (FSTs), as they are rule-based which allows language experts to define how the finite-state machine should behave depending on the language. As a result, FSTs make it possible to lemmatize words and produce mini- and full-paradigms. In this work, we use Jack Rueter's FSTs for Skolt Sami \cite{rueter-hamalainen-2020-fst}, Erzya and Moksha \cite{rueter-etal-2020-open}, and Komi-Zyrian \cite{71debe3f0608471284bc2419a2f392de}. The FSTs are supplied as part of the  UralicNLP~\cite{uralicnlp_2019} Python library. 

\subsection{Word embeddings of resource-rich languages}
Word embeddings are a vector representation of words, which are built based on the surrounding context of the word. Semantic similarity between words captured in the word embeddings can be measured using cosine similarity, which can then be utilized to cluster meanings in text~\cite{letsfaceit}. Common usages for word embeddings is to acquire semantically similar words to an input word. For example, the most 5 similar words to ``king'' are ``queen'', ``monarch'', ``prince'', ``sultan'', and ``ruler''. The vector nature of these words makes it possible to perform vector operations such as addition, multiplication and subtractions. With such operations, analogies could be predicted such as ``king'' - ``man'' + ``woman'' = ``queen''. Simply, this asks what is the equivalent of a king that is not a man but rather a woman in the semantic space, the answer is a queen.

When building word embeddings, there are many preprocessing configurations and hyperparameters that influence the performance of the models, such as lemmatization, part-of-speech tagging, window size, the dimension size of the embeddings, minimum and maximum thresholds for word frequencies and so on. There is no fixed nor optimal configuration that is apt for all applications.

In the translation dictionaries, words and their translations are provided in their lemma form. Due to this reason, the vocabulary in any word embeddings we will be using has to be lemmatized. Ideally, all the hyperparameters and configurations for word embeddings should be the same to capture similar features and semantics, which would yield better results across models once they are aligned. For the scope of this research, we use the most similar models we could get our hands on. 

We utilize the Russian and English~\cite{Velldal}, and Finish~\cite{laippala2014syntactic} word embeddings. The Russian embeddings are trained on a news corpus, while the English is based on Wikipedia and Gigaword 5th Edition corpora~\cite{parker2011english}. The Finnish word embeddings are trained on Common Crawls. The dimension size of the English and Russian embeddings is 300 but 200 is the size of the Finnish one. The window size is 5 for all embeddings but Finnish, which is 2. These differences, in addition to other reasons, end up affecting the quality of the models we will build of endangered languages. We discuss them more in the Discussion section.



\section{Cross-lingual word embeddings for endangered languages}
Cross-lingual word embeddings are word embeddings where vectors across multiple languages are aligned. For instance, the vector for ``dog'' in the English embeddings points roughly to the same direction for the same word in other languages (i.e., ``koira'' and ``\foreignlanguage{russian}{собака}'' for Finnish and Russian, respectively). Example applications for employing cross-lingual word embeddings are: headline generation~\cite{notime}, loan word identification \cite{mi-etal-2018-toward} and cognate identification \cite{lefever-etal-2020-identifying}.

Before we build and align the word embeddings, we apply a dimensionality reduction using the method proposed in~\cite{raunak-etal-2019-effective} to the three pre-trained models (i.e., English, Russian and Finnish). We set the target dimension to 100. This is to ensure that the vectors in all the embeddings share the same size. Subsequently, we process the vocabulary of the Finnish by removing all occurrences of the hashtag symbol ``\#'', which is there to mark compounds. Regarding the Russian word embeddings, the vocabulary contained part-of-speech information and, hence, each lemma might be present multiple times. To address this, we discard the part-of-speech information and use all vectors matching the target lemma. 

\begin{figure}
\includegraphics[width=\textwidth]{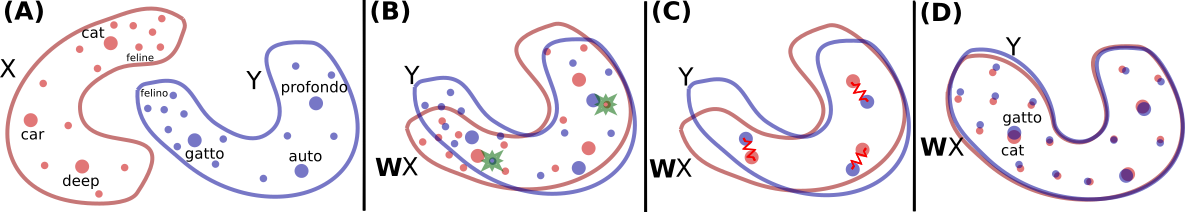}
\caption{A visualization of the transformation process of aligning word embeddings in $X$ in accordance to the ones in $Y$, taken from~\cite{MUSE}.} \label{fig:muse}
\end{figure}

To align the main three word embedding models, we employ the state-of-the-art supervised multilingual word embeddings  alignment technique introduced in MUSE~\cite{MUSE}. Figure~\ref{fig:muse} illustrates transforming the word embeddings of the source language $X$ with the target language $Y$ so that words in both languages are aligned together. In this example the source language is English and the target language is Italian. What supervised means in this context is that the alignment process relies on a bilingual dictionary that guides the transformation process. In our work, we set the target language to English and align both Russian and Finnish models with it using the bilingual dictionaries released as part of MUSE. The models are refined over 20 iterations.

Following the alignment of the resource-rich models, we construct the word embeddings for the endangered languages: Erzya, Moksha, Komi-Zyrian and Skolt Sami. In doing so, we iterate over all the lexemes in the dictionary of a given endangered language. In the case where a lexeme had translations to any of the three resource-rich languages and the translation existed in the word embeddings of the corresponding language, a vector for the lexeme is constructed as the centroid --an average vector-- of all translation vectors.

Once the word embeddings for the endangered languages have been constructed, we fine-tune them using the sentences in their universal dependencies. Lastly, we realign each word embeddings model with the resource-rich language having most translations to. In other words, Erzya and Skolt Sami are aligned with Finnish but Komi-Zyrian and Moksha are aligned with Russian and English, respectively. The models are aligned over 5 refinement steps.

\begin{table*}
\caption{The top 3 semantically similar words to the English vector, in all languages. The score after the colon is the semantic similarity score (the higher, the more similar).}\label{tab:top-sim}
\centering
\begin{adjustbox}{max width=\textheight, angle=90}
\centering
\begin{tabular}{|c|l|l|l|l|l|l|}
\hline
\textbf{eng (input)} & \textbf{fin}	&	\textbf{rus}	& \textbf{myv}	&	\textbf{mdf}	&	\textbf{sms}	&	\textbf{kpv}	 \\
\hline
dog	&	koira: 0.7100	&	\foreignlanguage{russian}{поймать}: 0.4354	&	\foreignlanguage{russian}{пинелевкс}: 0.8172	&	\foreignlanguage{russian}{пине}: 0.9977	&	piânnai: 0.7197	&	\foreignlanguage{russian}{гиджгысь}: 0.5691 \\
	&	kissa: 0.6618	&	\foreignlanguage{russian}{убивать}: 0.4310	&	\foreignlanguage{russian}{псарня}: 0.6691	&	\foreignlanguage{russian}{кутерь}: 0.8220	&	piânn\textcrg{}až: 0.7078	&	\foreignlanguage{russian}{барсук}: 0.5271 \\
	&	namipalo: 0.6263	&	\foreignlanguage{russian}{.родственник}: 0.4271	&	\foreignlanguage{russian}{киска}: 0.6340	&	\foreignlanguage{russian}{ката}: 0.7547	&	kaazzâž: 0.6521	&	\foreignlanguage{russian}{вежель}: 0.5238 \\
\hline
cat	&	rotta: 0.6631	&	\foreignlanguage{russian}{щенок}: 0.5246	&	\foreignlanguage{russian}{обизьган}: 0.7474	&	\foreignlanguage{russian}{ката}: 0.9990	&	kaazzâž: 0.6800	&	\foreignlanguage{russian}{черепаха}: 0.6121 \\
	&	Fretti: 0.6484	&	\foreignlanguage{russian}{бася}: 0.4885	&	\foreignlanguage{russian}{лыркай}: 0.7018	&	\foreignlanguage{russian}{зака}: 0.9990	&	žee$\textsuperscript{{$\prime$}}$vet: 0.6672	&	\foreignlanguage{russian}{питон}: 0.5996 \\
	&	kissa: 0.6461	&	\foreignlanguage{russian}{детеныша}: 0.4794	&	\foreignlanguage{russian}{пинелевкс}: 0.6993	&	\foreignlanguage{russian}{пине}: 0.7478	&	kue$\textsuperscript{{$\prime$}}$ttžeevai: 0.6665	&	\foreignlanguage{russian}{ӧбезьяна}: 0.5937 \\
\hline
king	&	Ahasveros: 0.5551	&	\foreignlanguage{russian}{наследник}: 0.4433	&	\foreignlanguage{russian}{инеазормастор}: 0.6751	&	\foreignlanguage{russian}{кароль}: 0.9971	&	daam: 0.5301	&	\foreignlanguage{russian}{королева}: 0.4869 \\
	&	kuningas: 0.5522	&	\foreignlanguage{russian}{гордо}: 0.4188	&	\foreignlanguage{russian}{озадкс}: 0.6643	&	\foreignlanguage{russian}{тюштя}: 0.8601	&	koon\textcrg{}õs: 0.5214	&	\foreignlanguage{russian}{принц}: 0.4736 \\
	&	kuninkas: 0.5243	&	\foreignlanguage{russian}{исходемик}: 0.4122	&	\foreignlanguage{russian}{инеазоронь}: 0.6315	&	\foreignlanguage{russian}{оцязор}: 0.7768	&	lââ$\textsuperscript{{$\prime$}}$ssnõmm: 0.5035	&	\foreignlanguage{russian}{герцог}: 0.4648 \\
\hline
queen	&	kuningatar: 0.5902	&	\foreignlanguage{russian}{паркер-боулз}: 0.5578	&	\foreignlanguage{russian}{инеазорава}: 0.9954	&	\foreignlanguage{russian}{оцязорава}: 0.9972	&	koon\textcrg{}õskaav: 0.7180	&	\foreignlanguage{russian}{королева}: 0.7227 \\
	&	prinsessa: 0.5867	&	\foreignlanguage{russian}{энистон}: 0.5314	&	\foreignlanguage{russian}{инеазоронь}: 0.5428	&	\foreignlanguage{russian}{kevärj}: 0.6191	&	prince$\textsuperscript{{$\prime$}}$ss: 0.5865	&	\foreignlanguage{russian}{принцесса}: 0.6614 \\
	&	kruununprinsessa: 0.5686	&	\foreignlanguage{russian}{чад}: 0.5063	&	\foreignlanguage{russian}{венчакай}: 0.5360	&	\foreignlanguage{russian}{лемдяз}: 0.6191	&	kå$\textsuperscript{{$\prime$}}$ll-lå$\textsuperscript{{$\prime$}}$dd: 0.5457	&	\foreignlanguage{russian}{принц}: 0.4903 \\
\hline
car	&	auto: 0.7728	&	\foreignlanguage{russian}{машина}: 0.6621	&	\foreignlanguage{russian}{автомобиль}: 0.7716	&	\foreignlanguage{russian}{машина}: 0.8568	&	autt: 0.6826	&	\foreignlanguage{russian}{мотик}: 0.6299 \\
	&	kottero: 0.6843	&	\foreignlanguage{russian}{бмв}: 0.6234	&	\foreignlanguage{russian}{автомашина}: 0.7716	&	\foreignlanguage{russian}{автокрандаз}: 0.6957	&	mõõnnâmneävv: 0.6572	&	\foreignlanguage{russian}{водитель}: 0.5915 \\
	&	katumaasturi: 0.6627	&	\foreignlanguage{russian}{bmw}: 0.6170	&	\foreignlanguage{russian}{уаз}: 0.7438	&	\foreignlanguage{russian}{ардомбяль}: 0.6377	&	luâđastvuejjamautt: 0.6438	&	\foreignlanguage{russian}{автобусса}: 0.5914 \\
\hline
man	&	spolle: 0.5062	&	\foreignlanguage{russian}{пожизненно}: 0.4450	&	\foreignlanguage{russian}{муюкт}: 0.4911	&	\foreignlanguage{russian}{аля}: 0.7974	&	so$\textsuperscript{{$\prime$}}$rmmjeei: 0.4548	&	\foreignlanguage{russian}{айулов}: 0.4970 \\
	&	pedofiiliä: 0.5029	&	\foreignlanguage{russian}{остин::пауэрс}: 0.4377	&	\foreignlanguage{russian}{нарт}: 0.4911	&	\foreignlanguage{russian}{ава}: 0.5362	&	upsee$\textsuperscript{{$\prime$}}$r: 0.4522	&	\foreignlanguage{russian}{катаржик}: 0.4538 \\
	&	puukottaja: 0.4986	&	\foreignlanguage{russian}{кривенко}: 0.4291	&	\foreignlanguage{russian}{гурямка}: 0.4869	&	\foreignlanguage{russian}{сакал}: 0.5212	&	nuõrrooumaž: 0.4468	&	\foreignlanguage{russian}{допроситны}: 0.4063 \\
\hline
woman	&	romaninainen: 0.5813	&	\foreignlanguage{russian}{юлия::печерская}: 0.5349	&	\foreignlanguage{russian}{аваломань}: 0.5539	&	\foreignlanguage{russian}{ава}: 0.9988	&	neezzan: 0.6134	&	\foreignlanguage{russian}{айулов}: 0.4585 \\
	&	somalinainen: 0.5713	&	\foreignlanguage{russian}{столбова\&mdash}: 0.5157	&	\foreignlanguage{russian}{авасыме}: 0.5428	&	\foreignlanguage{russian}{авакань}: 0.6255	&	ååumai: 0.4610	&	\foreignlanguage{russian}{мам}: 0.4035 \\
	&	maahanmuuttajanainen: 0.5436	&	\foreignlanguage{russian}{воспитатель}: 0.5079	&	\foreignlanguage{russian}{пекиязь}: 0.4938	&	\foreignlanguage{russian}{ни}: 0.5704	&	åålm: 0.4610	&	\foreignlanguage{russian}{колготки}: 0.3830 \\
\hline
France	&	Ranska: 0.6330	&	\foreignlanguage{russian}{франция}: 0.5325	&	\foreignlanguage{russian}{французонь}: 0.4922	&	\foreignlanguage{russian}{Кранцмастор}: 0.9964	&	Franskkjânnam: 0.6357	&	\foreignlanguage{russian}{забастовка}: 0.4077 \\
	&	Belgia: 0.6097	&	\foreignlanguage{russian}{деша}: 0.4916	&	\foreignlanguage{russian}{француз}: 0.4922	&	\foreignlanguage{russian}{кранц}: 0.7155	&	Jõnn-Britann: 0.5778	&	\foreignlanguage{russian}{кӧрень}: 0.3972 \\
	&	Iso-Britannia: 0.5757	&	\foreignlanguage{russian}{арно}: 0.4801	&	\foreignlanguage{russian}{австриец}: 0.4586	&	\foreignlanguage{russian}{кранцава}: 0.7155	&	Itaal: 0.5331	&	\foreignlanguage{russian}{японец}: 0.3698 \\
\hline
Finland	&	Tanska: 0.5735	&	\foreignlanguage{russian}{тудегешев}: 0.4599	&	\foreignlanguage{russian}{Финляндия}: 0.4457	&	\foreignlanguage{russian}{шведонь}: 0.6399	&	Lä$\textsuperscript{{$\prime$}}$dd: 0.6780	&	\foreignlanguage{russian}{ненеч}: 0.4165 \\
	&	Norja: 0.5732	&	\foreignlanguage{russian}{инсбрук}: 0.4462	&	\foreignlanguage{russian}{Суоми}: 0.4457	&	\foreignlanguage{russian}{шведава}: 0.6399	&	Lää$\textsuperscript{{$\prime$}}$ddjânnam: 0.6780	&	\foreignlanguage{russian}{вужкыв}: 0.3451 \\
	&	Viro: 0.5612	&	\foreignlanguage{russian}{либерец}: 0.4398	&	\foreignlanguage{russian}{Россия}: 0.4384	&	\foreignlanguage{russian}{швед}: 0.6399	&	Taarr: 0.6424	&	\foreignlanguage{russian}{подувкыв}: 0.3451 \\
\hline
see	&	Muuttu: 0.4886	&	\foreignlanguage{russian}{видеть}: 0.5243	&	\foreignlanguage{russian}{покш}: 0.5228	&	\foreignlanguage{russian}{няемс}: 0.9982	&	õinn: 0.5315	&	\foreignlanguage{russian}{дзик}: 0.4642 \\
	&	tämä: 0.4860	&	\foreignlanguage{russian}{тренд}: 0.5057	&	\foreignlanguage{russian}{те}: 0.5200	&	\foreignlanguage{russian}{няема}: 0.5860	&	o$\textsuperscript{{$\prime$}}$ddjõõttâd: 0.4936	&	\foreignlanguage{russian}{эсся}: 0.4590 \\
	&	ainavain: 0.4824	&	\foreignlanguage{russian}{тенденция}: 0.4935	&	\foreignlanguage{russian}{но}: 0.4920	&	\foreignlanguage{russian}{ила-крда}: 0.5248	&	tiett-aa: 0.4881	&	\foreignlanguage{russian}{шензьӧдлыны}: 0.4540 \\
\hline
want	&	haluta: 0.5709	&	\foreignlanguage{russian}{жить}: 0.5960	&	\foreignlanguage{russian}{мирямс}: 0.5852	&	\foreignlanguage{russian}{пиштемс}: 0.6032	&	soovšed: 0.5562	&	\foreignlanguage{russian}{гажавны}: 0.8506 \\
	&	siksi: 0.5219	&	\foreignlanguage{russian}{хотеть}: 0.5225	&	\foreignlanguage{russian}{секс}: 0.5654	&	\foreignlanguage{russian}{мезевок}: 0.5864	&	haa$\textsuperscript{{$\prime$}}$leed: 0.5319	&	\foreignlanguage{russian}{желайтны}: 0.6250 \\
	&	molempi: 0.5146	&	\foreignlanguage{russian}{актриса::юлия::михалков}: 0.5222	&	\foreignlanguage{russian}{одямс}: 0.5654	&	\foreignlanguage{russian}{мезе-мезе}: 0.5861	&	jee$\textsuperscript{{$\prime$}}$res: 0.5257	&	\foreignlanguage{russian}{ньӧтчыдысь}: 0.5561 \\
\hline
day	&	lomaaamu: 0.5336	&	\foreignlanguage{russian}{утра}: 0.4900	&	\foreignlanguage{russian}{поздаямс}: 0.4964	&	\foreignlanguage{russian}{цяс}: 0.5621	&	minut: 0.4695	&	\foreignlanguage{russian}{мӧдасув}: 0.4533 \\
	&	reissupäivä: 0.5230	&	\foreignlanguage{russian}{7days.ru}: 0.4613	&	\foreignlanguage{russian}{покшнэ}: 0.4948	&	\foreignlanguage{russian}{ой}: 0.5621	&	jâđđa: 0.4684	&	\foreignlanguage{russian}{вежонпом}: 0.4377 \\
	&	lähtöa: 0.5088	&	\foreignlanguage{russian}{выплакать}: 0.4506	&	\foreignlanguage{russian}{час}: 0.4730	&	\foreignlanguage{russian}{шиньгучка}: 0.5351	&	kõskkpei$\textsuperscript{{$\prime$}}$vv: 0.4673	&	\foreignlanguage{russian}{салют}: 0.3980 \\
\hline
\end{tabular}
\end{adjustbox}
\end{table*}

\section{Sentiment analysis}

In this section, we describe an experiment with the newly produced word embeddings. We apply them in the task of sentiment analysis. We hand pick all positive and negative sentences from the Erzya treebank \cite{rueter-tyers-2018-towards} based on the translations provided in the treebank in English and Finnish. This constitutes our Erzya test corpus that contains 23 negative sentences and 22 positive sentences, giving us a total of 45 sentences.  

We use the Stanford Sentiment Treebank for English \cite{socher-etal-2013-recursive} to train our sentiment analyzer model. As the Erzya test data is binary -- negative and positive sentences -- we treat the sentiment information in the treebank as binary as well, ignoring any neutral examples. It is important to note that we do not use any examples written in Erzya during the training, only sentences in English.

We train a neural model that takes in a sentence in English as a source and a sentiment label (positive or negative) as a target. We train the neural model with the aligned embeddings by substituting the words in the input sentences with their vectors. As our models are lemmatized, we need to ensure that all words are lemmatized in the in input as well. We use spaCy \cite{spacy} for this lemmatization step. The architecture and training of the neural model is inspired by the work presented in~\cite{joulin-etal-2017-bag}, where bi-grams are added to the input sentences during the training phase and the neural network is a linear classifier.


\begin{table}
\caption{Example sentences in Erzya and their translations in English, along with the predicted sentiment by our method for each sentence.}\label{tab:sentiment-examples}
\centering
\resizebox{\textwidth}{!}{%
\begin{tabular}{|l|l|c|}
\hline
\textbf{Erzya} & \textbf{English} & \textbf{Sentiment} \\
\hline
\foreignlanguage{russian}{Зярошкаль цёрыненть кенярксозо!}  &  You can imagine the boy's delight!  &  \multirow{3}{*}{Positive} \\\cline{1-2}
\foreignlanguage{russian}{Чизэ лембе.}  &  It is a warm day.  &   \\\cline{1-2}
\foreignlanguage{russian}{Сехте паро шка.}  &  The best time of all.  &  \\\hline
\foreignlanguage{russian}{Цёрынентень аламодо визькс теевсь.}  &  The boy felt a little ashamed.  &  \multirow{3}{*}{Negative} \\\cline{1-2}
\foreignlanguage{russian}{Баягинень ёмавтомась — пек берянь тешксэсь.}  &  Losing a bell was a really bad sign.  &  \\\cline{1-2}
\foreignlanguage{russian}{Весе те — апаро вийтнень тандавтнемс.}  &  This is all meant to scare away the evil spirits.  &  \\\hline
\end{tabular}
}
\end{table}

Table \ref{tab:sentiment-examples} shows some examples of the input in Erzya, its translation in English and the correctly predicted label. For Erzya, we use the lemmas from the treebank, and get their closest English vectors through the aligned word embeddings. This way, the model treats the Erzya sentences as though they were English and it can predict the sentiment in the language it did not see during the training. The resulting model was trained for 30 epochs and it reached to 53.3\% accuracy for Erzya and 75.5\% accuracy for English in the treebank sentences and an accuracy of 83.5\% in English in the Stanford Sentiment Treebank dataset. We have obtained an accuracy boost for Erzya predictions, reaching 57.8\%, when we also considered vectors of other resource-rich languages with the aid of the translation dictionary (Finnish in this case, as Erzya has many translations to Finnish).

The resulting accuracy is respectable given that the test data is fundamentally different from the training data. First of all, the testing and training are in different languages. Second of all, they represent very different genres: the training data is based on movie reviews, whereas the testing data has sentences from novels.


\section{Discussion and Conclusions}

The work conducted in this paper has been a first step for using machine learning in modelling the semantics of some of the endangered Uralic languages. It is evident that these aligning based approaches embraced before in the literature cannot get us too far in truly representing the semantics do to socio-cultural mismatches in concepts. For instance, we saw that \textit{Finland}, which is a very important concept for a Finnish model was completely misaligned with geographically close countries such as \textit{Denmark}, \textit{Norway} and \textit{Estonia}. Alignment can only get us so far and using models trained on larger languages has its inherent problems when applied to completely new domains in a completely different language.

Even the starting quality for the pretrained embeddings was low. The Russian model was unacceptably bad and the Finnish model has too many words that are not lemmatized at all, or are lemmatized to a wrong lemma. When the quality of the models available for a high-resourced languages is substandard, one cannot expect any sophisticated machine learning method to come to the rescue. Unfortunately in our field, too little attention is paid to the quality of resources and more attention is paid into single values representing overall accuracies and overall performance.

As there is no shortcut to happiness, we should look into the data available in the endangered languages themselves. For instance, FU-Lab has a plethora of resources for Komi languages \cite{fedina_komi_corpus,fedina_corpus_komi} that are just waiting for lemmatization. Once lemmatized, these resources could be used to build word embeddings directly in that language. Of course, this requires collaboration between many parties and willingness to make data openly available. While this might not be an issue with FU-Lab, it might be with some other instances holding onto their immaterial rights too tight.

At the current, stage our dictionary editing system, Ve$\textsuperscript{{$\prime$}}$rdd \cite{verdd_demo,verdd_2}, contains words for multiple endangered languages and their translations in a graph structure. This data could be extended by predicting new relations into the graph with semantic models such as word embeddings. This could help at least in resolving meaning groups and polysemy of the lexical entries. However, the word embeddings available for the endangered languages in question has not yet reached to a stage mature enough for their incorporation as a part of the lexicon.

\section{Acknowledgement}
I would like to dedicate the acknowledgement section to Jack Rueter, for all his work on endangered languages and his brilliant ideas on improving the current technological state of endangered languages. Jack's enthusiasm and dedication to endangered languages is clearly shown in all the various dictionaries and FSTs built and maintained by him. He supervised my work on building the dictionary editing system, Ve$\textsuperscript{{$\prime$}}$rdd \cite{verdd_demo,verdd_2}. He was always available for discussing and supporting my work, without him Ve$\textsuperscript{{$\prime$}}$rdd would not be in the great level it is at at the moment. He truly is a pioneer in the field, and the entire community appreciates all of his work. 

\bibliographystyle{splncs04}
\bibliography{bibliography}
\end{document}